\documentclass[letterpaper, 10 pt, conference]{ieeeconf}  % Comment this line out if you need a4paper

\IEEEoverridecommandlockouts% This command is only needed if 
\usepackage{url}
\usepackage{xcolor}
\usepackage{multirow}
\usepackage{adjustbox}
\usepackage{gensymb}
\usepackage{graphics} % for pdf, bitmapped graphics files
\graphicspath{ {images} }
\usepackage{epsfig} % for postscript graphics files
\usepackage{mathptmx} % assumes new font selection scheme installed
\usepackage{times} % assumes new font selection scheme installed
\usepackage{amsmath} % assumes amsmath package installed
\usepackage{amssymb}  % assumes amsmath package installed
\usepackage[ruled]{algorithm2e}
\usepackage[backend=bibtex,style=ieee]{biblatex} %CRD added backend=bibtex for arXiv version
\usepackage{hyperref}
\usepackage[]{quoting}
\usepackage{caption} 
\usepackage{subcaption}
\usepackage{newtxtext, newtxmath} % try to fix mathcal for sets!

\DeclareMathOperator*{\argmin}{arg\,min}
\newcommand\R{\mathbb{R}}

\title{\LARGE \bf
Risk-Conditioned Distributional Soft Actor-Critic \\ % CRD wrapped line
for Risk-Sensitive Navigation
}

\author{Jinyoung Choi$^1$,\hspace{-0.5mm} Christopher Dance$^2$,\hspace{-0.5mm} Jung-eun Kim$^1$,\hspace{-0.5mm} Seulbin Hwang$^1$,\hspace{-0.5mm} Kyung-sik Park$^1$ % <-this % stops a space
\thanks{$^1$NAVER LABS, Gyeonggi-do, 13494, South Korea}
\thanks{$^2$NAVER LABS Europe, 6 chemin de Maupertuis, Meylan, 38240, France.
Website: \url{europe.naverlabs.com}}
\thanks{$^1$jy-choi@naverlabs.com}
\thanks{$^2$chris.dance@naverlabs.com\vspace{3mm}} % CRD added vspace
\thanks{\copyright $\,$ 2021 IEEE. Personal use of this material is permitted. Permission from IEEE must be obtained for all other uses, in any current or future media, including reprinting/republishing this material for advertising or promotional purposes, creating new collective works, for resale or redistribution to servers or lists, or reuse of any copyrighted component of this work in other works.}
}

\begin{document}

\maketitle
\thispagestyle{empty}
\pagestyle{empty}

%%%%%%%%%%%%%%%%%%%%%%%%%%%%%%%%%%%%%%%%%%%%%%%%%%%%%%%%%%%%%%%%%%%%%%%%%%%%%%%%
\begin{abstract}

Modern navigation algorithms based on deep reinforcement learning (RL) 
show promising efficiency and robustness.
However, most deep RL algorithms operate in a risk-neutral manner, making no special attempt to shield users from 
relatively rare but serious outcomes,
even if such shielding might cause little loss of performance. Furthermore, such algorithms typically make no provisions to ensure safety in the presence of inaccuracies in the models on which they were trained, beyond adding a cost-of-collision and some domain randomization while training, in spite of the formidable complexity of the environments in which they operate.
In this paper, we present a novel distributional RL algorithm that not only learns an uncertainty-aware policy, but can also change its risk measure without expensive fine-tuning or retraining. Our method shows superior performance and safety over baselines in partially-observed navigation tasks. We also demonstrate that agents trained using our method can adapt their policies to a wide range of risk measures at run-time.

\end{abstract}

%%%%%%%%%%%%%%%%%%%%%%%%%%%%%%%%%%%%%%%%%%%%%%%%%%%%%%%%%%%%%%%%%%%%%%%%%%%%%%%%
\section{Introduction}

Deep reinforcement learning (RL) is attracting considerable interest in the field of mobile-robot navigation, due to its promise of superior performance and robustness compared with classical planning-based algorithms~\cite{faust2018prm,chiang2019learning}.
Despite this interest, few existing works on deep-RL-based navigation attempt to design risk-averse policies.
This is surprising for several reasons.
First, a navigating robot might cause harm to humans, to other robots, to itself or to its surroundings, and risk-averse policies may be safer than risk-neutral policies, while avoiding the over-conservative behaviour typical of policies based on worst-case analyses~\cite{majumdar2020}.
Second, in environments with such complex structure and dynamics that it is impractical to provide accurate models, policies optimizing certain risk measures are an appropriate choice, as they actually provide guarantees on robustness to modelling errors~\cite{chow2015}.
Third, the end-users, insurers and designers of navigation agents are risk-averse humans~\cite{kahneman2013}, so risk-averse policies seem to be the natural choice.

To address the issue of risk in RL, recent works~\cite{morimura2010, dabney2017distributional} have introduced the concept of \emph{distributional RL}. Distributional RL learns the distribution of accumulated rewards, rather than just the mean of that distribution. By applying an appropriate \emph{risk measure}, which is simply a mapping from this distribution of rewards to a real number, distributional RL algorithms can infer risk-averse or risk-seeking policies. 
Distributional RL 
has
shown superior sample efficiency and performance on arcade games~\cite{dabney2018implicit}, simulated robotics benchmarks~\cite{fan2020learning,Yue2020}, and real-world grasping tasks~\cite{bodnar2020}.
However, the theoretical basis for these excellent results is not understood~\cite{Lyle2019}, and it is not clear that these advantages would necessarily extend to navigation tasks. Moreover, one might prefer risk-averse policies in one environment, for instance to avoid scaring pedestrians, but such policies might be too risk-averse to pass through a narrow passage. So, one might need to train policies with different risk measures, suited to each environment, which would be computationally expensive and time-consuming.

In this paper, to efficiently train an agent that can adapt to multiple risk measures, we present the \emph{risk-conditioned distributional soft actor-critic} (RC-DSAC) algorithm which learns a wide range of risk-sensitive policies concurrently. In our experiments, RC-DSAC showed superior performance and safety compared with both non-distributional and distributional baselines. It could also adapt its policy to different risk measures without retraining. 

\begin{figure}[t]
\centering
\includegraphics[width=.75\columnwidth]{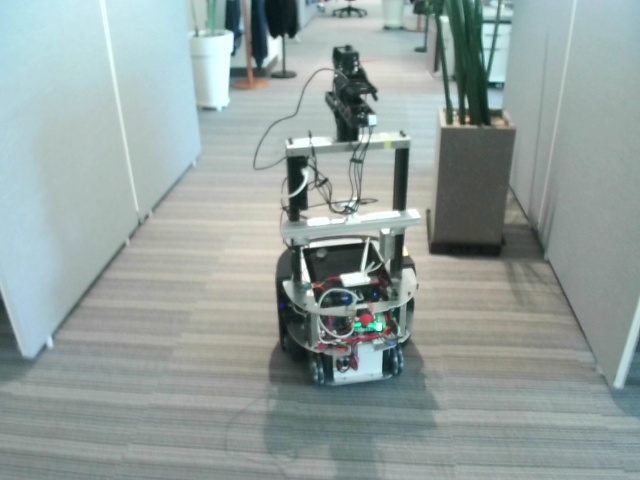} % CRD .25\textwidth -> .75\columnwidth
\caption{The robot and environment used in the real-world experiments of Section~\ref{section_experiments}.}
\label{figure_realworld}
%\vspace{-6mm} % CRD comment
\end{figure}

In summary, our main contributions are:
\begin{itemize}
\item A novel navigation algorithm based on distributional RL, that can learn a variety of risk-sensitive policies concurrently;
\item Improved performance over baselines in multiple simulation environments;
\item Generalization to a wide range of risk measures at run-time.
\end{itemize}

The next section discusses related work. Subsequent sections, explain our method and present experiments demonstrating its effectiveness. 

\section{Related Work}

\subsection{Risk in Mobile-Robot Navigation}
Although this paper takes a deep RL approach to safe and low-risk robot navigation, there is a vast literature on classical model-predictive-control (MPC) and graph-search approaches. 
This literature considers diverse sources of risk, ranging from simple sensor noise and occlusion \cite{Lunenburg2018,Kim2019}, to 
uncertainty about the traversability of the edges (e.g. doors) of a navigation graph \cite{Nardi2020}, and the unpredictability of pedestrian movements \cite{Mehta2018}. 

This literature has explored a wide variety of risk measures, ranging from collision probabilities \cite{Axelrod2018} used as chance constraints~\cite{Blackmore2011}, to entropic risk~\cite{Nishimura2020}. Interestingly,  \cite{Nishimura2020} took a hybrid approach, coupling deep learning for pedestrian motion prediction with nonlinear MPC, 
 arguing that only such a hybrid approach allows a robot's risk-metric parameters to be changed at run-time, unlike approaches relying on RL. To the contrary, the results of our paper demonstrate that such run-time parameter-tuning is straightforward for deep RL. 

While \cite{majumdar2020} recently argued that \emph{coherent risk measures} are well-suited to robotics, and we use coherent risk measures in this paper,
the only work that we are aware of that uses a coherent risk-metric coupled with MPC for robot navigation is \cite{Hakobyan2020}, which extends its authors' previous work \cite{Samuelson2018}, in which risk measures were applied to inventory control.

\subsection{Deep RL for Mobile-Robot Navigation} 
Deep RL has received much attention in the field of mobile-robot navigation due to its success in many game~\cite{espeholt2018impala,silver2017mastering} and robotics~\cite{haarnoja2018soft2,choi2020fast,akkaya2019solving,bodnar2020} domains.
Compared to classic approaches such as MPC, RL methods are known to be able to infer optimal actions without expensive trajectory predictions, and to perform more robustly when the cost or reward has local optima~\cite{long2017towards,chiang2019learning,chen2017socially}.

Recently, several deep-RL-based navigation methods have been proposed that explicitly account for risks arising from uncertainty about the environment.
As individual deep networks may make 
overconfident predictions on far-from-distribution samples,
\cite{lotjens2019safe} applied MC-dropout~\cite{gal2016dropout} and bootstrapping~\cite{osband2016deep} to predict collision probabilities.
An uncertainty-aware RL method was proposed in \cite{fan2020learning}, which has an additional observation-prediction model, and uses the prediction variance to adjust the variance of the actions taken by the policy. Meanwhile, \cite{kamran2020risk} designed `risk rewards' that encourage the safe behaviour of autonomous driving policies at lane intersections, and \cite{Katyal2020} proposed switching between two RL-based driving policies, based on the estimated uncertainty about future pedestrian motions.
Although these works show promising performance and improved safety in uncertain environments, they either require an additional prediction model, carefully shaped reward functions, or expensive Monte Carlo sampling at run-time. 

In contrast to existing works on RL-based navigation, we use \emph{distributional RL} to learn computationally-efficient risk-sensitive policies, without using an additional prediction model or a specifically-tuned reward function.

\subsection{Distributional RL and Risk-Sensitive Policies}

Distributional RL models the distribution of the accumulated reward, rather than just its mean.
While distributional RL was proposed over a decade ago \cite{morimura2010}, it received only scant mention in a comprehensive review of safe RL~\cite{Garcia2015} from 2015, and it has only been widely studied since its integration with deep learning~\cite{dabney2018implicit,barth2018distributed}.
Existing distributional RL algorithms rely on the recursion
\begin{gather}
\label{eq_distributional_Bellman}
Z^{\pi}(s,a) \overset{D}= r(s,a) + \gamma \,  Z^{\pi}(S',A'),
\end{gather}
where the \emph{random return} $Z^{\pi}(s,a)$ is defined as the discounted sum of rewards when starting in state $s$ and taking action $a$ under policy $\pi$, the notation $A \overset{D}= B$ indicates that the random variables $A$ and $B$ have identical distributions, $r(s,a)$ is the random reward given the state-action pair, $\gamma \in [0, 1)$ is the discount factor, the random state $S'$ follows the transition distribution given $(s,a)$, and the random action $A'$ is drawn from policy $\pi$ in state $S'$.

Empirically, distributional RL algorithms have shown superior performance and sample efficiency in many game domains~\cite{bellemare2017distributional,dabney2017distributional}.
It has been argued that this is because predicting quantiles serves as an auxiliary task that enhances representation learning, but as yet there is little supporting evidence for this conjecture~\cite{Lyle2019}.

Of central importance to this paper, is the fact that distributional RL facilitates the learning of risk-sensitive policies. To extract a risk-sensitive policy, \cite{dabney2018implicit, yang2019fully} learned to predict arbitrary quantiles of the distribution of the random return, and select risk-sensitive actions by estimating various `distortion risk measures' by sampling quantiles. As such sampling must be performed for each potential action, this approach is not possible for continuous action spaces. So instead, \cite{madsac} recently combined the soft actor-critic (SAC) framework~\cite{haarnoja2018soft2} with distributional RL,  achieving a new state-of-the-art in risk-sensitive control tasks. 
In robotics, \cite{Singh2020} considered a sample-based distributional policy gradient algorithm and demonstrated improved robustness to actuation noise on OpenAI Gym tasks when using coherent risk measures. Meanwhile, \cite{bodnar2020} proposed the use of distributional RL to learn risk-sensitive policies for grasping tasks, showing superior performance over non-distributional baselines on real-world grasping data.

Despite the impressive performance demonstrated in these existing works, they are all limited to learning a policy for a \emph{single} risk measure at a time. This can be problematic since the desired risk measure can vary with the environment and situation.
So, in this work, we train a single policy that can adapt to a wide variety of risk measures.

\section{Approach}

In this section, we discuss the problem formulation and  proposed method in detail.

\subsection{Problem Formulation}

We consider a differential-wheeled robot navigating in two dimensions. 
The robot's shape is an octagon (Figure~\ref{figure_env_sshot}),
and its objective is to pass a sequence of waypoints without colliding with obstacles.

We formalize this problem as a partially-observed Markov decision process (POMDP) \cite{aastrom1965optimal}, with sets of states $\mathcal{S}^\text{PO}$, observations $\Omega$ and actions $\mathcal{A}$, a reward function $r:\mathcal{S}^\text{PO}\times \mathcal{A}\rightarrow \mathbb{R}$, and distributions for the initial state, for state $s_{t+1}\in \mathcal{S}^\text{PO}$ given state-action $(s_t, a_t) \in \mathcal{S}^\text{PO} \times \mathcal{A}$ and for observation $o_t \in \Omega$ given $(s_t, a_t)$.
As is typical when applying RL, we treat this POMDP as a Markov decision process (MDP)
with set of states $\mathcal{S}$ given by the episode-histories of the POMDP:  
\begin{align*}
\mathcal{S} = \{ (o_0, a_0, o_1, a_1 \dots, o_T) : o_t \in \Omega, a_t \in \mathcal{A}, T \in \mathbb{Z}_{\ge 0} \}.
\end{align*}
The MDP has the same action space $\mathcal{A}$ as the POMDP, and its reward, initial-state and transition distributions are those implicitly defined by the POMDP. Note that the  reward is random variable for the MDP, even though we define it as a function for the POMDP.

\subsubsection{States and Observations}

The full state, which is a member of the set $\mathcal{S}^\text{PO}$,
is the location of all waypoints, coupled with the locations, velocities and accelerations of all obstacles, but real-world agents only sense a fraction of this state. So in this work, an observation
\begin{gather*}
(o_\text{rng}, o_{\text{waypoint}}, o_{\text{velocity}}) \in \R^{180}
\times \R^6 \times \R^4 =: \Omega
\end{gather*}
consists of range-sensor measurements describing the location of nearby obstacles, measurements of the robot's location relative of the next two waypoints, and information about the robot's velocity.
Specifically, we define
\begin{gather*}
o_{\text{rng},i} = \mathbb{I}\{d_i \in (0.01, 3) \, \mathrm{m} \} (2.5+\log_{10}d_i) ,
\end{gather*}
where $\mathbb{I}\{\cdot\}$ is the indicator function, $d_i$ is the distance in meters to the nearest obstacle in the angular range $[2i-2,2i)$ degrees, relative to the $x$-axis of the robot's coordinate frame, 
and we set $o_{\text{rng},i}=0$ 
if there are no obstacles in the given direction. 
The waypoint observation is of the form
\begin{gather*}
o_{\text{waypoint}}=[
 \log_{10}{\delta_1}, \cos{\theta_1} ,\sin{\theta_1}, \log_{10}{\delta_2}, \cos{\theta_2}, \sin{\theta_2}],
\end{gather*}
where $\delta_1, \delta_2$ are the distances to the next waypoint and the waypoint after that, clipped to $[0.01, 100] \, \mathrm{m}$, and $\theta_1, \theta_2$ are the angles of those waypoints relative to the robot's $x$-axis.
Lastly, the velocity observation
$o_{\text{velocity}}=[v_{\text{c}},\omega_{\text{c}},v_{\text{u}},\omega_{\text{u}}]$
consists of the robot's current linear and angular velocities
$v_{\text{c}}$, $\omega_{\text{c}}$, and the desired linear and angular velocities $v_{\text{u}}$, $\omega_{\text{u}}$ calculated from the agent's previous action.

\subsubsection{Actions}

We use normalized two-dimensional vectors $u = (u_0, u_1) \in [-1,1]^2 =: \mathcal{A}$ as actions, in terms of which the desired linear and angular velocity of the robot are
\begin{align*}
v_{\text{u}} &=  w_{\text{minv}} (1-u_0)/2 + w_{\text{maxv}} (1+u_0)/2, \\
\omega_{\text{u}} &= \mathbb{I}\{ |w_{\text{max}\omega} u_1| \ge 15\, \mathrm{deg/s}\} \,
      w_{\text{max}\omega} u_1 ,
\end{align*}
where $w_{\text{minv}}=-0.2\,\mathrm{m/s}$, $w_{\text{maxv}}=0.6\,\mathrm{m/s}$, $w_{\text{max}\omega}=90\,\mathrm{deg/s}$.

These desired velocities are sent to the robot's motor controller, which clips them to the ranges $[v_\text{c}-w_{\text{accv}} \Delta t, v_\text{c}+\omega_{\text{accv}} \Delta t]$ and $[\omega_\text{c}-w_{\text{acc}\omega} \Delta t, \omega_\text{c} + w_{\text{acc}\omega}\Delta t]$ for maximum accelerations  $w_{\text{accv}}=1.5\,\mathrm{m/s^2}$ and $w_{\text{acc}\omega}=120\,\mathrm{deg/s^2}$,
where $\Delta t=0.02\,\mathrm{s}$ is the control period of the motor controller.
The agent's control period is larger than $\Delta t$, being uniformly sampled from $\{0.12, 0.14, 0.16\} \, \mathrm{s}$ when an episode begins in the simulation, and $0.15 \, \mathrm{s}$ in real-world experiments.

\subsubsection{Reward}

Our reward function encourages the agent to follow the waypoints efficiently, while avoiding collisions. 
Omitting dependence on the state and action for brevity, the reward has the form
$$r = r_{\text{base}} + r_{\text{goal}} + r_{\text{waypoint}} \cdot r_{\text{angular}} + r_{\text{coll}}.$$
The base reward $r_{\text{base}} = -0.02$ is given at every step, to penalize the agent for the time taken to reach the goal (the last waypoint), and $r_{\text{goal}} = 10$ is given when the distance between the agent and the goal is less then $0.15\,\mathrm{m}$. 

The waypoint reward is 
$$
r_{\text{waypoint}} = \max\{ -0.1,  \max\{0, v_c\} \cos{\theta_1} \},$$
where $\theta_1$ is the angle of the next waypoint relative to the robot's $x$-axis and $v_c$ is the current linear velocity. We set $r_{\text{waypoint}}$ to zero when the agent is in contact with an obstacle.

Reward $r_\text{angular}$ encourages navigation in straight lines, 
\begin{gather*}
 r_\text{angular} = 
       \begin{cases}
       1.2 & \text{if } |\omega_u| < 15\, \mathrm{deg/s} \\
       \max\left\{0.5, 1-\frac{|\omega_u|}{(120\, \mathrm{deg/s})} \right\} & \text{otherwise,} \\
 \end{cases}  
\end{gather*}
and $r_{\text{coll}} = -10$ is given if the agent collides with an obstacle.

\subsubsection{Risk-Sensitive Objective}
As in (\ref{eq_distributional_Bellman}), let $Z^\pi(s,a)$ be the \emph{random return} given by
\begin{align}
\label{eq_accumulated_reward}
Z^\pi(s,a) = \sum_{t=0}^\infty\gamma^t r(S_t, A_t),
\end{align}
where $(S_t, A_t)_{t\in \mathbb{Z}_{\ge 0}}$ is the random state-action sequence with  
$(S_0, A_0) = (s,a)$, 
given by the MDP's transition distribution and policy $\pi$, and $\gamma \in [0,1)$ is the discount factor. 

There are two main approaches to defining risk-sensitive decisions. Either one defines a \emph{utility function} $U: \R \rightarrow \R$, as in~\cite{Morgenstern1953} and selects an action $a$ that maximizes $\mathbb{E} \, U(Z^\pi(s,a))$ when in state $s$. Alternatively, as in~\cite{Yaari1987,dabney2018implicit} one 
considers the \emph{quantile function} of $Z^\pi$, defined by
$Z^\pi_\tau(s,a) := \inf\{ z \in \R : \mathbb{P}(Z^\pi(s,a) \le z) \ge \tau\}$ 
for \emph{quantile fraction} $\tau \in [0,1]$.
Then one defines a \emph{distortion function}, which is a mapping $\psi : [0, 1] \rightarrow [0,1]$ from quantile fractions to quantile fractions, and selects an action $a$ that maximizes the \emph{distortion risk measure} 
$\mathbb{E}_{\tau \sim U([0,1])} Z^\pi_{\psi(\tau)}(s,a)$ 
when in state $s$. 

In this work, we focus on two distortion risk measures, each with a scalar parameter $\beta$ that we call the \emph{risk-measure parameter}. 
The first is the widely-used~\cite{chow2015,dabney2018implicit,madsac} conditional value-at-risk (CVaR), which is the expectation of the fraction $\beta$ of least-favourable 
random returns,
and corresponds to the distortion function
\begin{align*}
\hspace{9mm} \psi^\text{CVaR}(\tau;\beta) &:= \beta\tau &&\text{for $\beta \in (0,1]$.}
\intertext{Lower $\beta$ results in a more 
risk-averse
policy and $\beta=1$ gives a  risk-neutral policy. 
The second is the power-law risk measure, given by the distortion function}
\hspace{9mm} \psi^\text{pow}(\tau;\beta) &:= 1-(1-\tau)^{1/(1-\beta)} &&\text{for $\beta<0$,}
\end{align*}
motivated by its good performance in grasping experiments~\cite{bodnar2020}. 
For the given parameter ranges, both  risk measures are coherent
in the sense of~\cite{Artzner1999}.

\subsection{Risk-Conditioned Distributional Soft Actor-Critic}

To efficiently learn a wide range of risk-sensitive policies, 
we propose the \emph{risk-conditioned distributional soft actor-critic} (RC-DSAC) algorithm. 

\subsubsection{Soft Actor-Critic Algorithm}

Our algorithm is based on the \emph{soft actor-critic} (SAC) algorithm~\cite{haarnoja2018soft2}, the term `soft' indicating \emph{entropy-regularized}.
SAC maximizes the accumulated rewards and the entropy of the policy jointly: 
\begin{gather}
    J(\pi)=\mathbb{E}_\pi \Big[ \sum_{t=0}^{\infty}\gamma^{t}[r(s_t,a_t) + \alpha\, H(\pi(\cdot|s_t))] \Big],
    \label{equation_sac_obj}
\end{gather}
where the expectation is over state-action sequences given by the policy $\pi$ and transition distribution, $\alpha \in \R_{\ge 0}$ is a temperature parameter which trades-off the optimization of reward and entropy, and $H(p(\cdot)) := -\mathbb{E}_{a \sim p(a)} \log p(a)$ denotes the entropy of a  distribution over actions which is assumed to have a probability density $p(\cdot)$. 

SAC has a \emph{critic} network that learns a soft state-action value function $Q^\pi : \mathcal{S} \times \mathcal{A} \rightarrow \R$, using the soft Bellman operator
\begin{align*}
    & T^\pi Q^\pi(s_t,a_t) := \mathbb{E}_\pi \big[ \big.
    \\ &\hspace{0.3cm}  
    r(s_t,a_t) + \gamma\, \left( Q^\pi(s_{t+1},a_{t+1}) - \alpha\,  \log\pi(a_{t+1}|s_{t+1}) \right) \, \big| \, s_t, a_t\big],
\end{align*}
and an \emph{actor} network that minimizes the Kullback-Leibler divergence between the policy and a distribution given by the exponential of the soft value function,
\begin{align*}
    \pi_\text{new} &= \argmin_{\pi' \in \Pi} \mathbb{E}_{s \sim \mathcal{D}^{\pi_\text{old}}} \Bigg[  D_\text{KL}\bigg( \pi'(\cdot|s)\ \bigg\Vert \ \frac{
    e^{\frac{Q^{\pi_\text{old}}(s,\cdot)}{\alpha}}
    }{Z^{\pi_\text{old}}_\text{part.}(s)} \bigg) \Bigg],
\end{align*}
where $\Pi$ is the set of policies that can represented by the actor network, 
$\mathcal{D}^\pi$ is the distribution over states induced by policy $\pi$ and the transition distribution, which is approximated in practice by experience replay,
and $Z^{\pi_\text{old}}_\text{part.}(s_t)$ is the partition function normalizing the distribution.

In practice, the reparameterization trick is often used. In that case, SAC samples actions as $a_t = f(s_t, \epsilon_t)$ where $f(\cdot, \cdot)$ is the mapping implemented by the actor network, and $\epsilon_t$ is a sample from a fixed distribution like a spherical Gaussian $\mathcal{N}$. 
The policy objective then has the form
\begin{align}
J(\pi)=
\mathbb{E}_{s \sim \mathcal{D}^\pi,\epsilon \sim \mathcal{N}}[Q(s,f(s,\epsilon))
 - \alpha \, \log\pi(f(s, \epsilon) | s)].
\label{equation_actor_loss}
\end{align}
For more details about SAC, we refer the reader to \cite{haarnoja2018soft,haarnoja2018soft2}.

\subsubsection{Distributional SAC and Risk-Sensitive Policies}

To capture the full distribution of accumulated rewards, rather than just its mean, \cite{madsac} recently proposed \emph{distributional SAC} (DSAC). As in  previous work on distributional RL \cite{dabney2017distributional,dabney2018implicit,yang2019fully}, DSAC uses quantile regression to learn this distribution. Such previous work is limited to finite action spaces, and DSAC overcomes this limitation using ideas from SAC.

Rather than using the random return $Z^\pi$ from  equation~(\ref{eq_accumulated_reward}), DSAC works with the \emph{soft random return} appearing in~(\ref{equation_sac_obj}), given by
\begin{align*}
Z^{\alpha,\pi}(s,a) &:= \sum_{t=0}^\infty \gamma^t  [r(S_t,A_t) - \alpha\, \log \pi(A_t|S_t))] ,
\end{align*}
where $(S_t,A_t)_{t\in\mathbb{Z}_{\ge 0}}$ is as in (\ref{eq_accumulated_reward}).

Like SAC, the DSAC algorithm has an actor and a critic. To train the critic, some quantile fractions $\tau_1, \dots, \tau_N$ and $\tau_1', \dots, \tau_{N'}'$ are sampled 
independently, and the critic minimizes the loss
\begin{gather}
L(s_t,a_t,r_t,s_{t+1}) = \frac{1}{N'} \sum_{i=1}^{N}\sum_{j=1}^{N'}\rho_{\tau_i}\big({\delta^{\tau_i,\tau_j'}_{t}}\big),
\label{equation_critic_loss}
\end{gather}
where for $x\in\R$, the quantile regression loss is 
\begin{align}
\label{eq_quantile_loss}
\rho_{\tau}(x) &= |\tau - \mathbb{I}\{ x < 0 \}| \, \min\{x^2, 2|x|-1\}/2,
\end{align}
and the temporal difference is
\begin{align*}
\delta^{\tau,\tau'}_t = r_t+\gamma [ {\hat{Z}_{\tau'}}'(s_{t+1},a_{t+1}) -\alpha\log\pi(a_{t+1}|s_{t+1})] - \hat{Z}_{\tau}(s_t,a_t),
\end{align*}
where $(s_t, a_t, r_t, s_{t+1})$ is a transition from the replay buffer,
$\hat{Z}_{\tau}(s,a)$ is the output of the critic, which is an estimate of the $\tau$-quantile of 
$Z^{\alpha,\pi}(s,a)$, and 
${\hat{Z}_{\tau'}}'(s,a)$ 
is the output of a delayed version of the critic known as the \emph{target critic}~\cite{haarnoja2018soft2}.

To train a risk-sensitive actor network, DSAC works with a 
distortion function $\psi$. 
Rather than direcly maximizing the corresponding distortion risk measure, DSAC substitutes $Q(s,a) =\hat{\mathbb{E}}_{\tau \sim U([0,1])} Z_{\psi(\tau)}(s,a)$ in equation (\ref{equation_actor_loss}), where $\hat{\mathbb{E}}$ denotes the average of a sample.

\subsubsection{Risk-Conditioned DSAC}

Although risk-sensitive policies learnt by DSAC show promising results~\cite{madsac} in multiple simulation environments, DSAC can learn only one type of risk-sensitive policy at a time. This may be problematic for mobile-robot navigation if the appropriate risk-measure parameter differs with the environment, and  users wish to tune it at run-time.

To address this issue, we propose the \emph{risk-conditioned distributional SAC} (RC-DSAC) algorithm, which extends DSAC to learn a wide range of risk-sensitive policies concurrently and can change 
its risk-measure parameter
without retraining.
RC-DSAC learns risk-adaptable policies for a distortion function $\psi(\cdot;\beta)$ with parameter $\beta$, by supplying $\beta$ as an input to the policy $\pi(\cdot | s, \beta)$, the critic $\hat{Z}_\tau(s,a; \beta)$,
and the target critic $\hat{Z}_{\tau'}'(s,a; \beta)$.
Specifically, the critic's objective (\ref{equation_critic_loss})  becomes
\begin{gather}
L(s_t,a_t,r_t,s_{t+1},\beta) = \frac{1}{N'} \sum_{i=1}^{N}\sum_{j=1}^{N'}\rho_{\tau_i}\big({\delta^{\tau_i,\tau_j',\beta}_{t}}\big),
\label{equation_critic_loss_ours}
\end{gather}
where $\rho_\tau(\cdot)$ is as in (\ref{eq_quantile_loss}) and the temporal difference is
\begin{multline*}
{\delta^{\tau,\tau',\beta}_t}=r_t+\gamma [ \hat{Z}'_{\tau'}(s_{t+1},a_{t+1};\beta)-\alpha\log\pi(a_{t+1}|s_{t+1},\beta)]\\ - \hat{Z}_{\tau}(s_t,a_t;\beta),
\end{multline*}
and the actor's objective (\ref{equation_actor_loss}) becomes
\begin{align}
\nonumber
J(\pi)&= 
\mathbb{E}_{s \sim \mathcal{D}^\pi,\epsilon \sim \mathcal{N}, \beta \sim \mathcal{B}} \big[\\
& \qquad Q(s,f(s,\epsilon, \beta);\beta) - \alpha \, \log\pi(f(s, \epsilon, \beta) | s) \big],
\label{equation_actor_loss_ours}
\end{align}
where $Q(s,a;\beta) =\hat{\mathbb{E}}_{\tau \sim U([0,1])}  {\hat Z}_{\psi(\tau;\beta)}(s,a;\beta)$ and $\mathcal{B}$ is a distribution for sampling $\beta$ as we now explain. 

During training, the risk-measure parameter $\beta$ is uniformly sampled from $\mathcal{B}=U([0,1])$ for $\psi^\text{CVaR}$, and $U([-2,0])$ for $\psi^\text{pow}$. As in other RL algorithms, each iteration has a \emph{data-collection} phase and a \emph{model-update} phase. In the data-collection phase, we sample $\beta$ at the start of each episode and fix it until the episode's end. For the model-update phase, we explore the following two alternatives. The first alternative, called \emph{stored}, stores the $\beta$ used in data-collection in the experience-replay buffer, and only uses that stored $\beta$ for updates. The second alternative, called \emph{resampling}, samples a new $\beta$ for each experience in a mini batch at every iteration. 

\subsubsection{Network Architectures}

\begin{figure}[t]
\vspace{3mm}
\centering 
\includegraphics[width=.75\columnwidth]{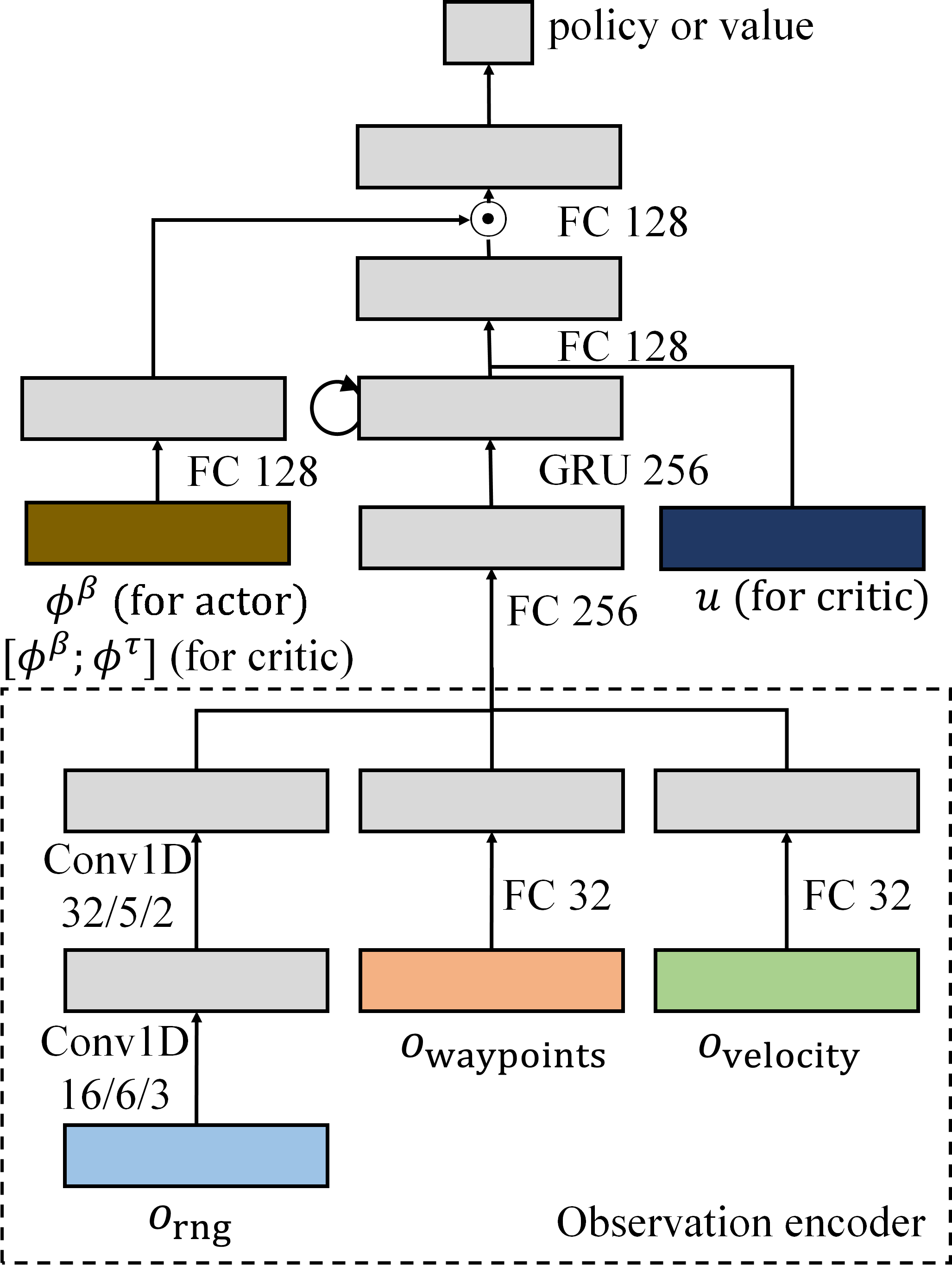} % CRD .25\textwidth -> 0.75\columnwidth
\caption{Architecture of the networks used in RC-DSAC. FC denotes a fully-connected layer, Conv1D denotes a one-dimensional convolutional layer with the given number of channels/kernel\_size/stride, and GRU denotes a gated recurrent unit \cite{cho2014properties}. Multiple arrows pointing to a single block indicate concatenation and $\odot$ denotes element-wise multiplication. }
\label{figure_architecture}
%\vspace{-5mm} % CRD comment
\end{figure}

We represent $\tau$ and $\beta$ using a cosine embedding, and use element-wise multiplication to fuse these with information about the observation and quantile fraction (Figure~\ref{figure_architecture}).

As in DSAC~\cite{madsac}, only the critic network of RC-DSAC depends on $\tau$. However, both the actor and critic networks of RC-DSAC depend on $\beta$.
So, we calculate embeddings $\phi^\beta \in \R^{64}$, $\phi^\tau \in \R^{64}$, with  elements $\phi^\beta_i = \cos(\pi i \beta)$ and $\phi^\tau_i = \cos(\pi i \tau)$.
Then, we apply element-wise multiplication
$$
g^\text{actor}(o_{0:t})\odot g^\text{actorRisk}(\phi^\beta)
$$
to the actor network and
$$
g^\text{critic}(o_{0:t},u_t)\odot g^\text{criticRisk}( [\phi^\beta;\, \phi^\tau] )
$$
to the critic network, where $g^\text{actor}(o_{0:t}),g^\text{critic}(o_{0:t},u_t) \in \R^{128}$ are the embeddings
of the observation history (and the current action for the critic) calculated using a gated recurrent unit (GRU) \cite{cho2014properties} and a fully-connected layer,  
$g^\text{actorRisk}:\R^{64} \rightarrow \R^{128}$ and $g^\text{criticRisk} : \R^{128} \rightarrow \R^{128}$ 
are fully-connected layers, and $[\phi^\beta;\, \phi^\tau]$ is the concatenation of vectors $\phi^\beta$ and $\phi^\tau$.

\section{Experiments}
\label{section_experiments}
In this section, we describe the simulation environment used for the training. Then we compare the performance of our method against baselines and demonstrate the trained policy using a real-world robot.

\subsection{Training Environment}

\begin{figure}[t]
\vspace{3mm}
\centering
    \begin{minipage}{0.6\linewidth}
      \begin{subfigure}{\linewidth}
        \centering
        \includegraphics[width=\textwidth]{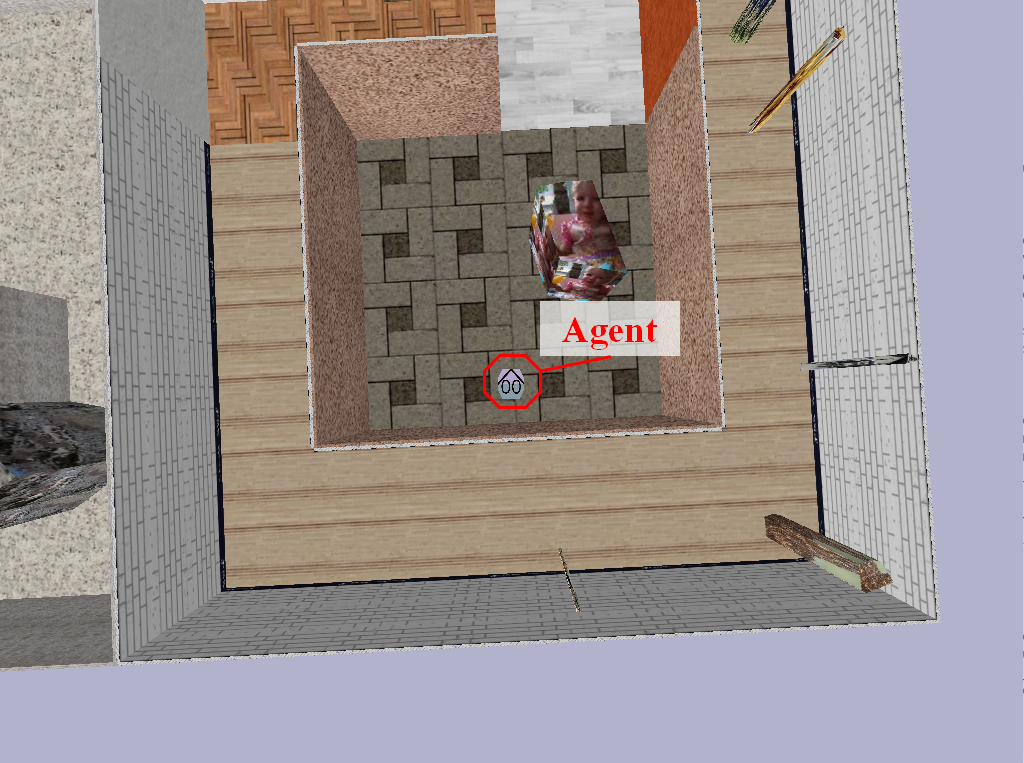}
        \caption{PyBullet simulator}
        \label{figure_env_sshot}
      \end{subfigure} 
    \end{minipage}
    \hfill
    \begin{minipage}{0.37\linewidth}
    \hfill
      \begin{subfigure}{0.5\textwidth}
        \includegraphics[width=\textwidth]{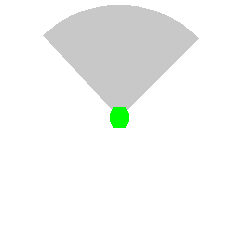}
        \vspace{-12mm}
        \caption{Narrow}
        \label{figure_env_narrow}
      \end{subfigure} 
    \hfill
    \\
    \hspace*{\fill}
      \begin{subfigure}{0.5\textwidth}
        \vspace{5mm}
        \includegraphics[width=\textwidth]{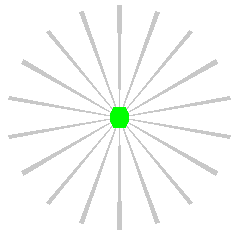}
        \caption{Sparse}
        \label{figure_env_sparse}
      \end{subfigure}
    \hfill
    \end{minipage}
    \hfill
\caption{(a) An example of our procedurally-generated environment. (b,c) Sensor settings: green octagons denote the robot and gray areas are the sensors' fields of view.}
\label{figure_env}
%\vspace{-5mm} CRD Comment
\end{figure}

We use the same procedural environment-generation algorithm as in~\cite{choi2020fast}, and use PyBullet~\cite{coumans2016pybullet} to simulate the robot dynamics, as shown in Figure~\ref{figure_env}~(a). 
To increase the throughput of data collection, we run 10 simulations in parallel. Specifically, for each environment generated, we run 10 episodes in parallel, where the episodes involve agents with distinct start and goal positions, as well as distinct risk-metric parameters $\beta$.
Each episode terminates after 1,000 steps, and a new goal is sampled when an agent reaches its goal.

To study the impact of partial observation on our method, we conduct experiments using two different sensor configurations, as shown in Figure~\ref{figure_env}~(b,c). The \emph{narrow} setting zeros out $o_{\text{rng},22:158}$, and the \emph{sparse} setting zeros out all $o_{\text{rng},i}$ except those with $i\equiv 0\ (\mathrm{mod}\ 10)$.

\subsection{Training Agents}\label{subsection_training_agents}

We compare the performance of RC-DSAC with SAC~\cite{haarnoja2018soft2} and DSAC~\cite{madsac}. We also compare with Algorithm~1 in~\cite{choi2020fast}, applied to this paper's reward function, calling this method \emph{reward-component-weight randomization} (RCWR).

We train two RC-DSAC agents, one for each of the distortion functions $\psi^\text{CVaR}$ and $\psi^\text{pow}$. Then RC-DSAC with $\psi^\text{CVaR}$ is evaluated for $\beta \in \{0.25,0.5,0.75,1\}$, and RC-DSAC with $\psi^\text{pow}$ is evaluated for $\beta \in \{-2,-1.5,-1,-0.5\}$.

For DSAC, we use $\psi^\text{CVaR}$ with $\beta \in \{0.25,0.75\}$, and $\psi^\text{pow}$ with $\beta \in \{-2,-1\}$, each DSAC agent being trained and evaluated for a single $\beta$.

For RCWR, we use only one \emph{navigation parameter}~\cite{choi2020fast} $w_\text{coll} \sim U([0.1,2])$.
The reward $r_\text{coll}$ is replaced by $w_\text{coll} r_\text{coll}$, when calculating the reward $r$, with higher values of $w_\text{coll}$ making an agent more collision-averse while still remaining risk-neutral. We use $w_\text{coll} \in \{1,1.5,2\}$ for evaluation.

All baselines use the same network architecture as RC-DSAC, with the following exceptions. DSAC does not use $g^\text{actorRisk}$, and $g^\text{criticRisk}$ depends only on $\phi^\tau$. RCWR has an extra  32-dimensional fully-connected layer in its observation encoder for $w_\text{coll}$. Lastly, RCWR and SAC use neither $g^\text{actorRisk}$ nor $g^\text{criticRisk}$.

We use the hyperparameters in Table~\ref{table_hyperparam} for all algorithms.
\begin{table}[h]
% \vspace{-5mm}
\caption{Hyperparameters}
\vspace{-1.5mm}
\label{table_hyperparam}
\centering
\begin{adjustbox}{max width=0.45\textwidth}
\begin{tabular}{|c|c|c|c|}
\hline
Parameter & Value & Parameter & Value \\ \hline
Learning rate & $3\times10^{-4}$ & \begin{tabular}[c]{@{}c@{}}Quantile fraction\\ samples ($N,N'$)\end{tabular} & 16 \\ \hline
Discount factor ($\gamma$) & 0.99 & \begin{tabular}[c]{@{}c@{}}Experience replay\\ buffer size\end{tabular} & $5\times10^6$ \\ \hline
\begin{tabular}[c]{@{}c@{}}Target network\\ update coefficient\end{tabular} & 0.001 & Mini-batch size & 100 \\ \hline
Entropy target~\cite{haarnoja2018soft2} & -2 & GRU unroll & 64 \\ \hline
\end{tabular}
\end{adjustbox}
\vspace{-2mm}
\end{table}

We train each algorithm for 100,000 weight updates (5,000 episodes in 500 environments). Then we evaluate the algorithms on 50 environments not seen in training. We evaluate for 10 episodes per environment, with 
agents having distinct start and goal positions, but having a common value for $\beta$ or $w_\text{coll}$.

To ensure fairness and reproducibility, we use fixed random seeds for training and evaluation, so different algorithms are trained and evaluated on exactly the same sequences of environments, and starting/goal positions.

\subsection{Performance Comparison}

\setlength{\tabcolsep}{2pt}
\begin{table}[t]
% \vspace{2mm}
\caption{Performance evaluation against baselines.}
\label{table_performance}
\newcommand{\xpm}[1]{{\tiny$\hspace{0.5mm} \pm\hspace{0.1mm}#1$}}
\centering\footnotesize
\begin{tabular}{|c|c|c|c|c|c|c|}
\hline
\multirow{2}{*}{Agent} & \multirow{2}{*}{$\psi$} & \multirow{2}{*}{$\beta$} & \multicolumn{2}{c|}{Narrow} & \multicolumn{2}{c|}{Sparse} \\ \cline{4-7} 
 &  &  & Collisions & Rewards & Collisions & Rewards \\ \hline
\multirow{8}{*}{\begin{tabular}[c]{@{}c@{}}RC-DSAC\\ (resample)\end{tabular}} 
 & \multirow{4}{*}{CVaR} & 0.25 & $0.67$\xpm{2.06} & $403.9$\xpm{186.2} & $0.19$\xpm{0.48} & $487.8$\xpm{\phantom{0}88.2} \\ \cline{3-7} 
 &  & 0.5 & $0.59$\xpm{1.03} & $451.3$\xpm{125.4} & $0.29$\xpm{0.62} & $512.0$\xpm{\phantom{0}54.8} \\ \cline{3-7} 
 & & 0.75 & $0.81$\xpm{1.75} & $452.0$\xpm{145.9} & $0.42$\xpm{0.93} & $507.6$\xpm{\phantom{0}65.1} \\ \cline{3-7} 
 &  & 1 & $1.15$\xpm{2.48} & $458.8$\xpm{140.3} & $0.55$\xpm{1.03} & $505.2$\xpm{\phantom{0}60.1} \\ \cline{2-7} 
 & \multirow{4}{*}{pow} & -2 & $0.50$\xpm{0.84} & $509.4$\xpm{\phantom{0}99.2} & $0.21$\xpm{0.68} & $473.4$\xpm{113.9} \\ \cline{3-7} 
 &  & -1.5 & $\mathbf{0.48}$\xpm{0.89} & $511.7$\xpm{\phantom{0}98.8} & $\mathbf{0.17}$\xpm{0.53} & $479.0$\xpm{107.4} \\ \cline{3-7} 
 &  & -1 & $0.58$\xpm{1.36} & $\mathbf{514.7}$\xpm{\phantom{0}96.4} & $0.21$\xpm{0.58} & $482.2$\xpm{101.9} \\ \cline{3-7} 
 &  & -0.5 & $0.68$\xpm{1.18} & $506.7$\xpm{113.3} & $0.23$\xpm{0.75} & $488.3$\xpm{104.2} \\ \hline
\multirow{8}{*}{\begin{tabular}[c]{@{}c@{}}RC-DSAC\\ (stored)\end{tabular}} & \multirow{4}{*}{CVaR} & 0.25 & $0.68$\xpm{3.47} & $443.5$\xpm{168.3} & $0.37$\xpm{0.68} & $494.7$\xpm{\phantom{0}89.3} \\ \cline{3-7} 
 &  & 0.5 & $1.00$\xpm{5.14} & $397.7$\xpm{173.2} & $0.38$\xpm{0.80} & $499.4$\xpm{\phantom{0}87.7} \\ \cline{3-7} 
 &  & 0.75 & $1.10$\xpm{2.27} & $431.0$\xpm{152.3} & $0.39$\xpm{0.77} & $501.0$\xpm{\phantom{0}86.0} \\ \cline{3-7} 
 & & 1 & $1.59$\xpm{8.09} & $298.4$\xpm{246.9} & $1.00$\xpm{1.63} & $477.7$\xpm{\phantom{0}97.6} \\ \cline{2-7} 
 & \multirow{4}{*}{pow}& -2 & $0.87$\xpm{3.90} & $465.0$\xpm{151.6} & $0.42$\xpm{0.72} & $492.3$\xpm{\phantom{0}84.5} \\ \cline{3-7} 
 &  & -1.5 & $0.73$\xpm{2.11} & $471.4$\xpm{130.0} & $0.68$\xpm{1.32} & $468.4$\xpm{335.8} \\ \cline{3-7} 
 &  & -1 & $1.13$\xpm{3.40} & $460.1$\xpm{122.2} & $0.58$\xpm{0.96} & $504.5$\xpm{\phantom{0}80.6} \\ \cline{3-7} 
 & & -0.5 & $0.95$\xpm{3.30} & $459.1$\xpm{122.9} & $1.12$\xpm{1.52} & $496.7$\xpm{\phantom{0}84.0} \\ \hline
\multirow{4}{*}{DSAC} & \multirow{2}{*}{CVaR} & 0.25 & $1.05$\xpm{1.75} & $431.9$\xpm{127.6} & $0.76$\xpm{1.18} & $417.2$\xpm{117.8} \\ \cline{3-7} 
 &  & 0.75 & $0.72$\xpm{3.00} & $299.6$\xpm{199.2} & $0.63$\xpm{1.03} & $515.4$\xpm{\phantom{0}74.1} \\ \cline{2-7} 
 & \multirow{2}{*}{pow} & -2 & $1.14$\xpm{4.02} & $469.2$\xpm{212.6} & $0.54$\xpm{1.29} & $\mathbf{525.5}$\xpm{\phantom{0}76.8} \\ \cline{3-7} 
 & & -1 & $0.73$\xpm{2.57} & $499.4$\xpm{115.7} & $0.80$\xpm{1.80} & $513.3$\xpm{\phantom{0}84.5} \\ \hline
\multirow{3}{*}{RCWR} & \multicolumn{2}{c|}{$w_\text{coll}=2$} & $1.58$\xpm{2.68} & $488.2$\xpm{122.5} & $0.81$\xpm{1.08} & $506.1$\xpm{\phantom{0}81.1} \\ \cline{2-7} 
 & \multicolumn{2}{c|}{$w_\text{coll}=1.5$} & $1.50$\xpm{2.39} & $491.7$\xpm{108.8} & $1.17$\xpm{1.71} & $491.9$\xpm{101.2} \\ \cline{2-7} 
 & \multicolumn{2}{c|}{$w_\text{coll}=1$} & $1.60$\xpm{2.55} & $493.7$\xpm{116.7} & $1.23$\xpm{1.59} & $490.8$\xpm{\phantom{0}93.5} \\ \hline
SAC & \multicolumn{2}{c|}{-} & $1.76$\xpm{2.02} & $476.7$\xpm{105.4} & $1.62$\xpm{2.48} & $491.8$\xpm{103.5} \\ \hline
\end{tabular}
\vspace{-3mm}
\end{table}

Table~\ref{table_performance} presents the mean and standard deviation of the number of collisions and reward of each method, averaged over the 500 episodes across the 50 evaluation environments.

RC-DSAC with $\psi^\text{pow}$ and $\beta=-1$ had the highest rewards in the narrow setting, and RC-DSAC with $\psi^\text{pow}$ and $\beta=-1.5$ had the fewest collisions in the both settings.

The risk-sensitive algorithms (DSAC, RC-DSAC) all had fewer collisions than SAC, and some of them could achieve this while attaining a higher reward.
Also, the results for RCWR suggest that distributional risk-aware approaches can be more effective than simply increasing the penalty for collisions.

We compare DSAC with the two alternative implementations of RC-DSAC by averaging  over both risk measures, but only for the two values of $\beta$ on which DSAC was evaluated.
In the narrow setting, RC-DSAC (stored) had 
a comparable number of 
collisions (0.95 vs. 0.91) but higher rewards (449.9 vs. 425.0) than DSAC, whereas in the sparse setting RC-DSAC (stored) had fewer collisions (0.44 vs. 0.68) 
but comparable %and higher 
rewards (498.1 vs. 492.9). Overall, RC-DSAC (resampling) had the fewest collisions (0.64 in the narrow setting and 0.26 in the sparse setting), and attained the highest rewards in the narrow setting (470.0). This shows the algorithm's ability to adapt to a wide range of risk-measure parameters, without the retraining required by DSAC.

In addition, the number of collisions made by RC-DSAC shows a clear positive correlation with $\beta$, for 
the CVaR risk measure. One would expect this, as low $\beta$ corresponds to risk aversion.

\subsection{Real-World Experiments}
To demonstrate the proposed method in the real world, we build a mobile-robot platform and test the agents trained in simulation (as described in Section~\ref{subsection_training_agents}), in a challenging office environment (Figure~\ref{figure_realworld}). The robot has four depth cameras on its front, and point cloud data from these sensors is mapped into the observation $o_\text{rng}$ corresponding to the narrow setting. Then we deploy RC-DSAC (resampling) and baseline agents. For each agent, we ran two experiments in a course of length $53.7\,\mathrm{m}$, making a run forward and another in the reverse direction.

\begin{table}[t]
\caption{Results of real-world experiment.}
\vspace{-3mm}
\label{table_real_world}
\begin{center}
\begin{adjustbox}{max width=0.45\textwidth}
\begin{tabular}{|c|c|c|c|c|c|c|}
\hline
\multirow{2}{*}{Agent} & \multirow{2}{*}{\begin{tabular}[c]{@{}c@{}}
$\psi$
\end{tabular}} & \multirow{2}{*}{$\beta$} & \multicolumn{2}{c|}{Forward} & \multicolumn{2}{c|}{Reverse} \\ \cline{4-7} 
 &  &  & Collision & Required Time ($\mathrm{s}$) & Collision & Required Time ($\mathrm{s}$) \\ \hline
\multirow{4}{*}{RC-DSAC} & \multirow{2}{*}{CVaR} & 0.25 & 0 & 107 & 0 & 114 \\ \cline{3-7} 
 &  & 0.75 & 0 & 112 & 1 & 109 \\ \cline{2-7} 
 & \multirow{2}{*}{pow} & -2 & 0 & 110 & 0 & 116 \\ \cline{3-7} 
 &  & -1 & 0 & 107 & 1 & 107 \\ \hline
\multirow{4}{*}{DSAC} & \multirow{2}{*}{CVaR} & 0.25 & 0 & 141 & 0 & 128 \\ \cline{3-7} 
 &  & 0.75 & 0 & 104 & 0 & 114 \\ \cline{2-7} 
 & \multirow{2}{*}{pow} & -2 & 0 & 109 & 0 & 104 \\ \cline{3-7} 
 &  & -1 & 0 & 111 & 0 & 104 \\ \hline
SAC & - & - & 3 & 115 & 2 & 111 \\ \hline
\end{tabular}
\end{adjustbox}
\end{center}
\vspace{-7mm}
\end{table}

Table~\ref{table_real_world} presents the number of collisions and required time to reach the goal for each agent. As can be seen, SAC had more collisions than distributional risk-averse agents. DSAC had no collisions throughout the experiments but showed over-conservative behaviour and took the longest time to reach the goal with $\psi^\text{CVaR}$ and $\beta=0.25$. RC-DSAC performed competitively with DSAC except minor collisions in less risk-averse modes, and could adapt its behaviour according to $\beta$.

We refer the supplementary video for the detailed trajectories and characteristics of each agent.

\section{Conclusion}

This paper proposed a novel distributional-RL method that produces agents that can adapt to a wide range of risk measures at run-time.
In our experiments, this method showed superior performance over the baselines, 
as well as 
adjustable risk-sensitivity. We also demonstrated the method using a real-world robot.

\section*{Acknowledgement}
We thank Tomi Silander and the NAVER LABS Robotics Group for their help with proofreading and the experiments.

\printbibliography

\end{document}